\documentclass{article}

\usepackage[preprint]{corl_2026} 

\usepackage[utf8]{inputenc} 
\usepackage[T1]{fontenc}    
\usepackage{hyperref}       
\usepackage{url}            
\usepackage{booktabs}       
\usepackage{arydshln}       
\usepackage{amsfonts}       
\usepackage{nicefrac}       
\usepackage{microtype}      
\usepackage{mathtools} 
\usepackage{amsmath} 
\usepackage{multirow}
\usepackage{bm}
\usepackage{svg}

\usepackage{todonotes}

\usepackage{float}
\usepackage{caption}
\usepackage{booktabs}
\usepackage{pifont}
\usepackage{adjustbox}
\newcommand{\cmark}{\ding{51}}
\newcommand{\xmark}{\ding{55}}

\usepackage{enumitem} 

\title{HapTile: A Haptic-Informed Vision-Tactile-Language-Action Dataset for Contact-Rich Imitation Learning}

\author{
Amirhosein Alian$^{1,\dagger}$,
Yongqiang Zhao$^{1,\dagger}$,
Shiyi Gu$^1$,
Xuyang Zhang$^1$,
Zhuo Chen$^1$,\\
{\bfseries
Christopher E. Mower$^{2}$,
Haitham Bou-Ammar$^{2,3}$,
Shan Luo}$^{1}$\\
$^1$King's College London, UK\\
$^2$Huawei, Noah's Ark Lab, UK\\
$^3$University College London, UK\\
Corresponding author: \texttt{shan.luo@kcl.ac.uk}
}

\def\ndatacollectors{9}
\def\ndemos{1,726}
\def\ntasks{38}
\def\nmanipulationskills{9}

\def\nminsdata{750.33}

\begin{document}

\maketitle
\setcounter{footnote}{0}
\renewcommand{\thefootnote}{\fnsymbol{footnote}}
\footnotetext{\textsuperscript{$\dagger$} Equal contribution.}

\begin{figure*}[h]
  \centering
  \includegraphics[width=1\textwidth]{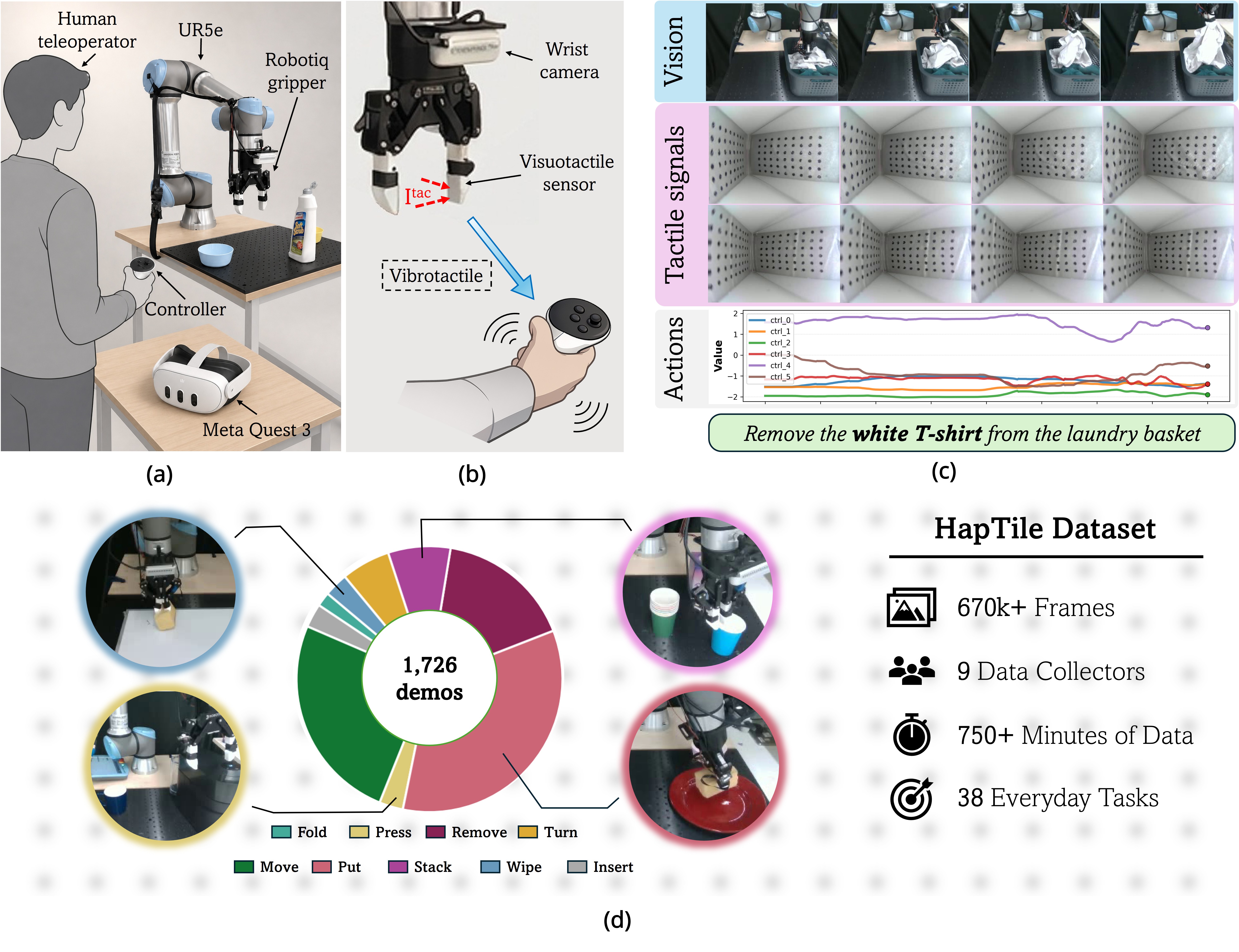} 
  \caption{HapTile Data collection architecture: (a) Experimental setup, (b) Haptic feedback from tactile variations, (c) Multimodal observations and actions, and (d) Dataset overview.
}
  \label{fig_intro}
\end{figure*}

\begin{abstract}
Despite the importance of tactile sensing for reliable manipulation, most existing Vision-Language-Action (VLA) datasets remain vision-only, and those that do incorporate tactile information typically lack the joint combination of task diversity, language conditioning, and action trajectories. Furthermore, existing teleoperation pipelines rarely provide haptic feedback to the operator, despite its established role in demonstration quality and manipulation stability. In this work, we present HapTile, a contact-grounded visuotactile manipulation dataset that advances beyond vision-only trajectory datasets by embedding physical interaction sensing at two levels: fingertip tactile feedback at the robot end-effector, and haptic-informed demonstrations at the teleoperator side. The data collection platform integrates haptic feedback directly into the teleoperation controller, enabling the operator to perceive contact interactions in real time. It is built around a standard and reproducible robotic system equipped with custom-designed fingertip tactile sensors. The dataset comprises everyday manipulation tasks spanning a broad range of contact-rich skills, including pick-and-place, folding, pressing, stacking, and other routine activities. Each task is paired with language instructions that condition the policy on the manipulation objective, together with synchronized visuotactile observations and action trajectories. In addition, we provide a benchmarking study on contact-rich policy learning using two baseline models to evaluate the effectiveness of the proposed contact-grounded dataset. The dataset and additional details are available on our website:  \href{https://haptile-dataset.github.io}{haptile-dataset.github.io}.
\end{abstract}

\section{Introduction}
\label{sec:intro}
Human manipulation relies on coordinated visual and tactile sensing~\cite{johansson2009coding}. Vision provides global information about object location and appearance, while tactile feedback captures localized contact properties such as slip, force, and surface deformation, particularly during compliant object manipulation~\cite{sun2025soft}. In tasks such as cloth handling and bottle opening, vision alone is often insufficient due to occlusions and limited contact visibility~\cite{lee2019making}. Tactile sensing complements vision by providing information invariant to lighting and viewpoint~\cite{wang2022spectac,luo2025tactile}, making both modalities essential for reliable everyday manipulation.

Vision-Language-Action (VLA) models have emerged as a rapidly growing direction in robotic manipulation. By fine-tuning large pre-trained vision-language models such as PaLI-Gemma~\cite{beyer2024paligemma} and CLIP~\cite{radford2021learning} to generate robot actions, these approaches enable robots to process visual observations and language instructions jointly. Recent models, including RT-2~\cite{zitkovich2023rt}, OpenVLA~\cite{kim2024openvla}, and $\pi_{0}$~\cite{pi0}, demonstrate strong performance across diverse manipulation tasks. However, they lack explicit tactile feedback required for contact-rich interactions. Training these models also requires large-scale demonstration data, further limited by the scarcity and poor reproducibility of contact-grounded datasets.

Recent work has incorporated tactile or force sensing into VLA models, including TLA~\cite{hao2025tla}, FuSe~\cite{jones2025beyond}, ForceVLA~\cite{yu2026forcevla}, Tactile-VLA~\cite{huang2025tactile}, 3D-ViTac~\cite{huang20243d}, and FD-VLA~\cite{zhao2026fd}. These studies demonstrate the importance of contact sensing for precise manipulation. However, existing tactile manipulation datasets are typically self-collected and limited in task diversity~\cite{bi2025vla,sliwowski2025demonstrating}. More broadly, no existing dataset jointly provides diverse everyday manipulation tasks with language conditioning, action trajectories, and embedded visuotactile feedback collected through haptic-informed teleoperation.

We present a contact-grounded manipulation dataset collected using a teleoperation platform with embedded vision-based tactile sensors. Unlike prior works~\cite{fu2024tvl,dave2024multimodal,fang2023rh20t,huang2026tafvla}, language instructions are used as policy conditioning signals rather than descriptive metadata. The dataset contains \ndemos\ demonstrations across \ntasks\ tasks and \nmanipulationskills\ manipulation skills, including pick-and-place, pressing, wiping, turning, folding, and compliant object handling. It includes objects with diverse mechanical and geometric properties, including items from the YCB object set~\cite{calli2015ycb}, and spans tasks with varying levels of tactile dependency. The platform is built on widely used hardware to ensure accessibility and reproducibility. Data collection incorporates real-time haptic feedback to the operator, improving demonstration quality. We further validate the dataset by training Diffusion Policy and $\pi_0$ baseline policies for manipulation learning. In summary, the contributions of our study are:



\begin{itemize}[leftmargin=*]

    \item A multi-modal, contact-grounded dataset of daily-life manipulation tasks is presented, collected using a universally deployable robotic platform. 
    
    \item Haptic feedback is embedded in the teleoperation pipeline and recorded in the dataset to improve demonstration quality and operative awareness. 
    \item A reproducible, low-cost vision-based tactile sensor compatible with industrial and open-source grippers is presented for large-scale data collection.
    \item Language instructions are used as policy conditioning signals rather than descriptive metadata. To the best of the authors’ knowledge, no existing dataset jointly provides language conditioning, tactile sensing, and robot action trajectories with haptic-informed demonstrations.

    \item The dataset and code will be open-sourced to enable reproducible data collection.

\end{itemize}

\begin{table*}[t]
\centering
\caption{Comparison of manipulation datasets. Source refers to the method used for data collection. F/T = 6-DOF Force/Torque; VBT = Vision-Based Tactile Sensor; PZR = Piezoresistive; GS = GelSight. \cmark$^{*}$ indicates the use of language as descriptive metadata rather than for policy conditioning. }
\label{tab:dataset_comparison}
\resizebox{\textwidth}{!}{%
\begin{tabular}{lcccccl c}
\toprule
\textbf{Dataset} & \textbf{Language} & \textbf{Action} & \textbf{Tactile} & \textbf{Haptic} & \textbf{Source} & \textbf{Sensor} \\
\midrule
RoboSet~\cite{bharadhwaj2024roboagent}      & \cmark            & \cmark & \xmark & \xmark   & Tele-op  & --          \\
DROID~\cite{khazatsky2024droid}             & \cmark$^{*}$      & \cmark & \xmark & \xmark   & Tele-op  & --          \\
RH20T~\cite{fang2023rh20t}                  & \cmark$^{*}$      & \cmark & \cmark & External & Tele-op  & Wrist F/T   \\
\hdashline
\noalign{\vskip 3pt}
Touch and Go~\cite{yang2022touch}           & \xmark            & \xmark & \cmark & \xmark   & Manual   & GS          \\
TVL~\cite{fu2024tvl}                        & \cmark$^{*}$      & \xmark & \cmark & \xmark   & Manual   & GS          \\
MMWand~\cite{dave2024multimodal}            & \cmark$^{*}$      & \xmark & \cmark & \xmark   & Manual   & F/T + VBT   \\
\hdashline
\noalign{\vskip 3pt}
VTDexManip~\cite{yu2025vtdexmanip}          & \xmark            & \xmark & \cmark & \xmark   & Manual   & PZR         \\
VTG~\cite{wen2024vtg}                       & \xmark            & \xmark & \cmark & \xmark   & Scripted & PZR         \\
REASSEMBLE~\cite{sliwowski2025demonstrating}   & \xmark            & \cmark & \cmark & \xmark   & Scripted & Wrist F/T   \\
TaF-VLA~\cite{huang2026tafvla}              & \cmark$^{*}$      & \cmark & \cmark & \xmark   & Tele-op  & VBT + F/T   \\
Touch in the Wild~\cite{zhu2025touch}       & \xmark            & \cmark & \cmark & \xmark   & Manual   & PZR         \\
\midrule
\textbf{HapTile (Ours)}                               & \cmark            & \cmark & \cmark & Embedded & Tele-op  & Custom VBT   \\
\bottomrule
\end{tabular}%
}
\end{table*}

\section{Related Work}
\label{sec:related}

\textbf{Large-Scale Robot Manipulation Datasets.} Robot generalization remains limited by data availability, motivating increasingly large and diverse datasets~\cite{o2024open,walke2023bridgedata,brohan2022rt,jang2022bc,gao2022objectfolder}. Bharadhwaj et al.~\cite{bharadhwaj2024roboagent} show that 7,500 curated demonstrations from RoboSet with semantic augmentation enable a single language-conditioned policy to perform 12 skills across 38 tasks. In contrast, DROID~\cite{khazatsky2024droid} aggregates 76k trajectories across 564 scenes from 18 laboratories on three continents, emphasizing scene diversity for zero-shot generalization. RH20T~\cite{fang2023rh20t} prioritizes multimodal richness, providing 110k sequences with visual, force, audio, and proprioceptive signals across 140 tasks. However, none of these datasets incorporates fingertip tactile sensing, except RH20T, which includes tactile measurements in only one configuration without extensive evaluation. The remaining datasets rely on wrist-level force sensing, capturing global contact loads rather than spatial contact patterns critical for slip detection, contact localization, grip adjustment, and operator haptic feedback.

\textbf{Visuotactile Datasets and Representation Learning.} Another line of work focuses on learning generalizable representations from visuotactile data collected outside robotic manipulation tasks. Touch and Go~\cite{yang2022touch} contains 14k tactile interactions across 4,000 object instances using a GelSight sensor~\cite{yuan2017gelsight} in diverse environments. Despite its diversity, it lacks manipulation goals and robot actions, limiting its use for policy learning. Fu et al.~\cite{fu2024tvl} extend this direction with the TVL dataset, pairing 44k visuotactile samples with GPT-4V-generated language annotations to train a tri-modal contrastive encoder across touch, vision, and language. Similarly, Chi et al.~\cite{dave2024multimodal} introduce MMWand, collected with a handheld device to align tactile, visual, and language modalities. While these datasets advance representation learning, they do not support direct policy learning due to the absence of task structure, goals, and action trajectories.

\textbf{Tactile Sensing in Contact-Rich Manipulation.} Other works integrate tactile sensing into specific manipulation tasks. VTDexManip~\cite{yu2025vtdexmanip} provides 2,032 human demonstrations across 10 daily tasks using a piezoresistive tactile glove, although human hand kinematics introduces a domain gap for robotic transfer. VTG~\cite{wen2024vtg} focuses on three-finger grasping across 18 objects but is limited to grasp stability. REASSEMBLE~\cite{sliwowski2025demonstrating} incorporates event cameras, wrist force-torque sensing, and audio, yet tactile sensing remains wrist-based and language annotations are absent. On the VLA side, TaF-VLA~\cite{huang2026tafvla} aligns tactile and force inputs in a shared latent space using a temporal contrastive adapter and over 10 million tactile-force frames, outperforming vision-only baselines on force-sensitive tasks. Touch in the Wild~\cite{zhu2025touch} introduces a portable tactile-enabled gripper with 2,700 demonstrations across 43 tasks in 12 environments, alongside a masked autoencoding framework for fine-grained manipulation. Despite these advances, no existing dataset jointly provides diverse everyday tasks, language annotations, action trajectories, and fingertip tactile sensing on an accessible robotic platform. A comparison is provided in Table~\ref{tab:dataset_comparison}.

\section{HapTile Dataset}
\label{sec:dataset}
\begin{figure*}[t]
  \centering
  \includegraphics[width=0.9\textwidth]{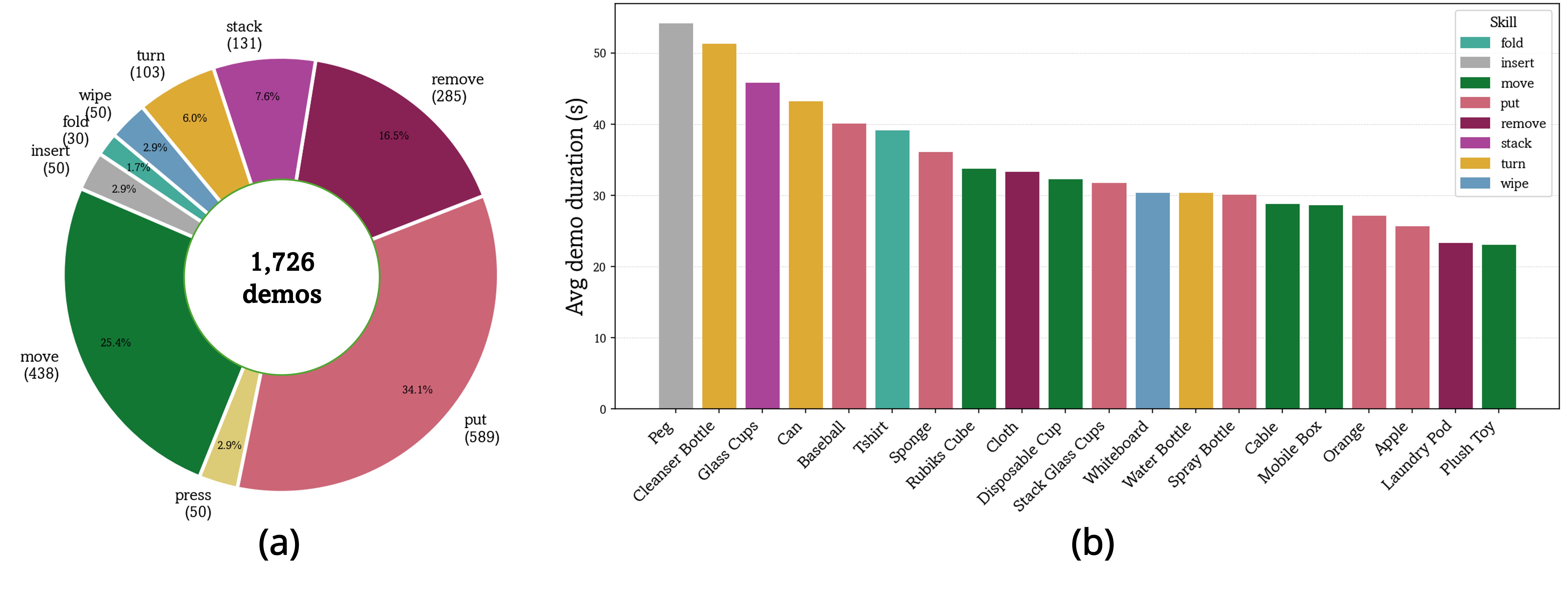} 
  \caption{HapTile dataset: (a) The diversity of the skills and their corresponding collected demonstrations, (b) Average demonstration data collection time per task.      
}
  \label{fig_statistics}
\end{figure*}

In this section, we introduce HapTile and describe how the dataset was collected, processed, and structured for policy learning. 
We first analyze demonstration statistics, then present 
synchronization, and quality-control procedures. Data formatting mechanisms are detailed in Section \ref{sec:appendix_formatting}. 

\subsection{Dataset Statistics}
HapTile is a visuotactile, contact-grounded manipulation dataset with \ndemos\ demonstrations across \ntasks\ tasks, collected by \ndatacollectors\ human operators via a haptic-informed teleoperation interface. It covers \nmanipulationskills\ manipulation skills with everyday objects, including YCB items~\cite{calli2015ycb}: \textit{move}, \textit{put}, \textit{remove}, \textit{turn}, \textit{fold}, \textit{insert}, \textit{wipe}, \textit{press}, and \textit{stack}, reflecting common daily interactions. Each demonstration includes language instructions and synchronized vision, tactile data, robot state, and action trajectories, sampled at 15~Hz. 
Fig.~\ref{fig_statistics}(a) shows the skill distribution, dominated by \textit{move}, \textit{put}, and \textit{remove}. The dataset totals \nminsdata\ minutes of interaction. Fig.~\ref{fig_statistics}(b) reports the top 20 tasks by duration, reflecting task difficulty, with the longest being inserting a peg into a hole (54.21~s) and the shortest moving a golf ball (12.42~s).


\subsection{Synchronization and Data Quality Control}

All data modalities are synchronized through the robot control loop. 
For policy learning, actions are converted to a unified 7D end-effector delta representation
\begin{equation}
    a_t =
    [\Delta x, \Delta y, \Delta z, \Delta r, \Delta p, \Delta \psi, g],
\end{equation}
where 
$\Delta x, \Delta y, \Delta z$ are translational deltas, 
$\Delta r, \Delta p, \Delta \psi$ are rotational deltas, and $g \in \left[0, 1\right]$ is the gripper command. 
This decouples learning from the exact robot configuration, enabling cross-embodiment by focusing the policy on local contact adjustment from tactile feedback.

Several quality checks are applied to every collected trajectory.
Empty or corrupted trajectories are removed, and 
timestamp gaps are inspected to detect dropped frames. 
Action-state consistency is checked to avoid confusing absolute commands with delta actions.  

\section{Data Collection Platform}
\label{sec:collectionsetup}
HapTile is constructed using a teleoperation system~\cite{lin2025learning} designed for contact-grounded manipulation rather than purely geometric trajectory recording. The system
captures third-person vision, wrist vision, bilateral tactile images, robot proprioception, language instructions, operator actions, and haptic feedback states.
 
\begin{figure}[t]
  \centering
  \includegraphics[width=0.8\textwidth]{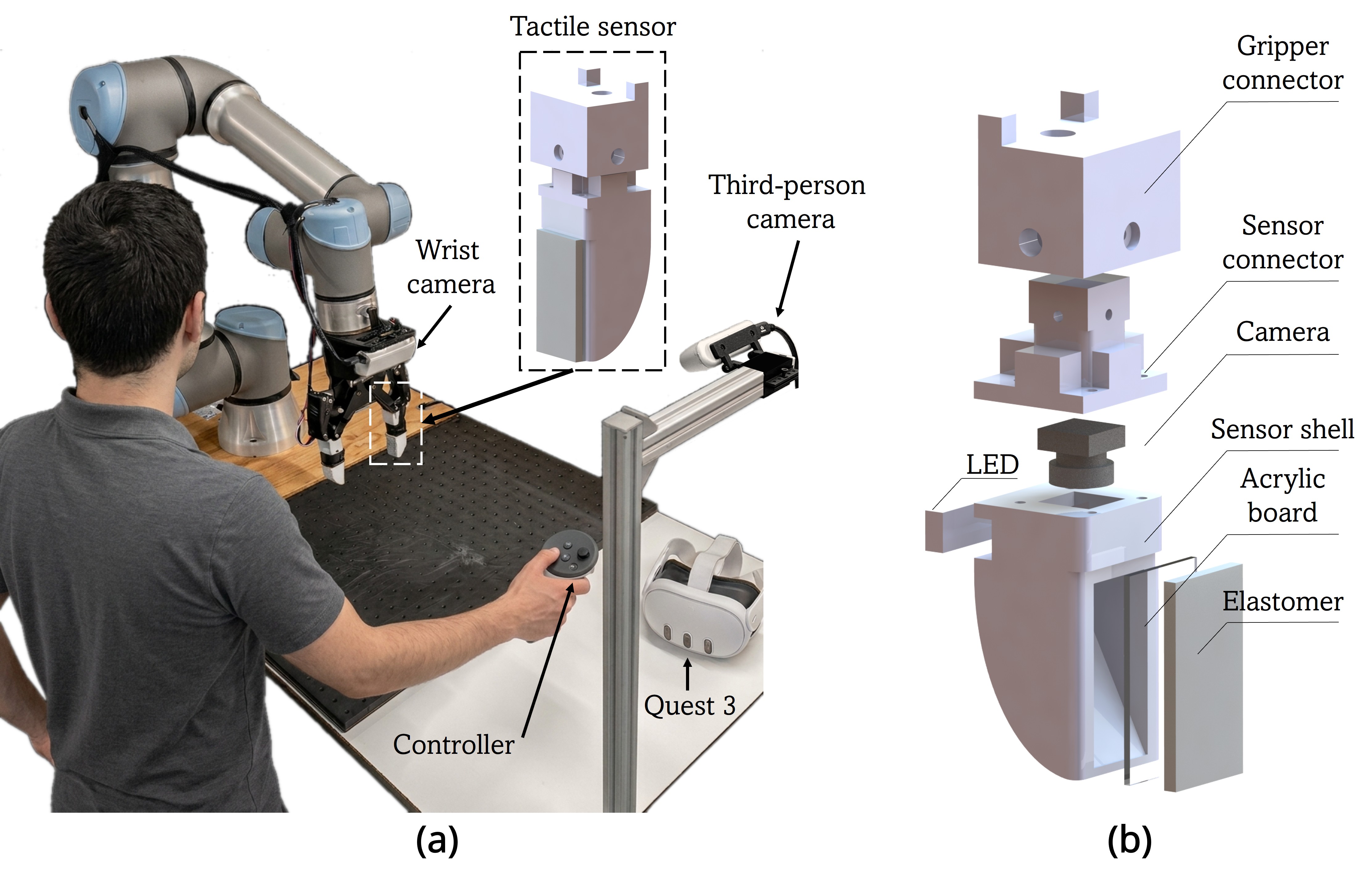} 
  \caption{Experimental setup to collect the HapTile dataset: (a) The overview of the hardware components, (b) The exploded view of the finger tip embedded with vision-based sensors.      
}
  \label{fig_setup}
\end{figure}

\subsection{System Overview}

Our data collection system, shown in Fig.~\ref{fig_setup}(a), consists of two components: a robot execution and sensing stack, and a teleoperator interface. 
The robot stack combines a UR5e manipulator, a Robotiq 2F-85 gripper with two custom fingertip vision-based tactile sensors (Fig.~\ref{fig_setup}(b)), wrist-mounted and third-person RGB cameras, and trajectory logging.
The teleoperator interface uses a Meta Quest headset/controller to command end-effector motion and gripper state, which are converted into end-effector pose targets.

At each control step, the system records an observation-action tuple:
\begin{equation}
    (o_t, a_t, l, \tau_t),
\end{equation}
where $o_t$ includes visual, tactile, and robot-state observations, $a_t$ is the expert action, $l$ is the task language instruction, and $\tau_t$ is the timestamp. 
The visual observations include a third-person camera for global scene context and a wrist camera to capture object-relative contact interactions. The tactile observations are captured from two vision-based tactile sensors mounted on the gripper fingers.


\subsection{Vision-Based Tactile Sensing and Marker Tracking}

Each gripper finger is equipped with a vision-based tactile sensor (Fig.~\ref{fig_setup}(b)) measuring approximately $20 \text{mm} \times 23 \text{mm} \times 37 \text{mm}$ with an effective sensing area of $16\text{mm} \times 29.5\text{mm}$. The sensing layer consists of a $3\text{mm}$ transparent silicone elastomer (XP-565, A:B=1:1) coated with a diffuse reflective silicone–gray ink mixture to enhance deformation visibility. An RGB camera (IMX258, 30 FPS, 135\textdegree~FOV) captures tactile images in real time under uniform illumination from an internal COB LED strip.

\begin{figure}[t]
  \centering
  \includegraphics[width=0.9\textwidth]{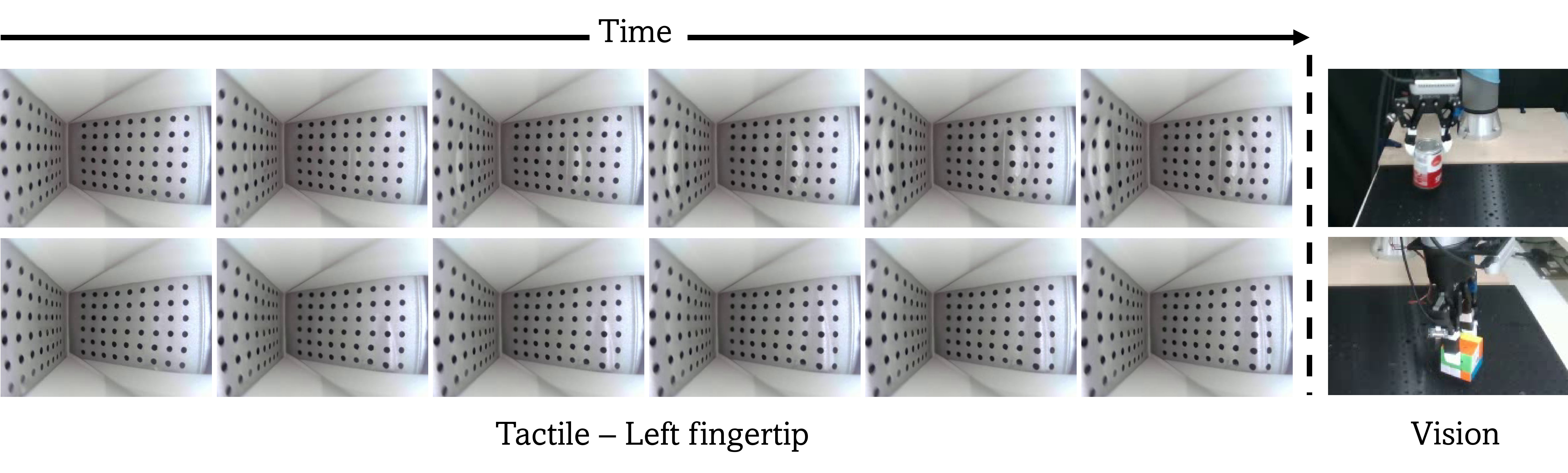} 
  \caption{Left visuotactile sensor outputs during can and Rubik’s cube manipulation (top to bottom), showing pre-grasp to grasp transitions with corresponding third-person views on the right.      
}
  \label{fig_rawtactile}
\end{figure}

As shown in Fig.~\ref{fig_rawtactile}, contact induces characteristic image changes and marker displacements that depend on the contact geometry and the  task. RGB tactile images from both fingers are recorded at each control step, and further processing of the raw tactile images is detailed in Section~\ref{sec:appendix_tactile}.




Our system estimates tactile surface motion online by tracking marker displacement in vision-based tactile sensor streams.
At the start of each trajectory, marker points are detected in the tactile image and used as references. These points are then tracked at each timestep using Lucas–Kanade optical flow~\cite{bouguet2001pyramidal}. To improve robustness, we apply forward–backward consistency checks, marker-region masking, and a minimum-valid-point threshold. If too few points remain valid, the reference frame is reinitialized.

We compute displacement relative to the initial reference for all valid markers. Global drift can optionally be compensated by subtracting the median displacement across markers, yielding a local displacement signal, which is then converted into a scalar motion score using

\begin{equation}\label{eq:tactile-motion-estimate}
    m_t = \sum_i \max \left( \lVert \Delta p_t^i \rVert_2 - \epsilon, 0 \right),
\end{equation}
where $\epsilon$ is a motion deadband. 
This score approximates contact-induced deformation and potential slip. 
We smooth the signal with separate rise and release factors, producing a stable tactile motion estimate. 
The resulting marker-motion signal is stored in HapTile and is also used to drive haptic feedback.

\subsection{Haptic Feedback to the Operator}

\begin{figure}[t]
  \centering
  \includegraphics[width=0.9\textwidth]{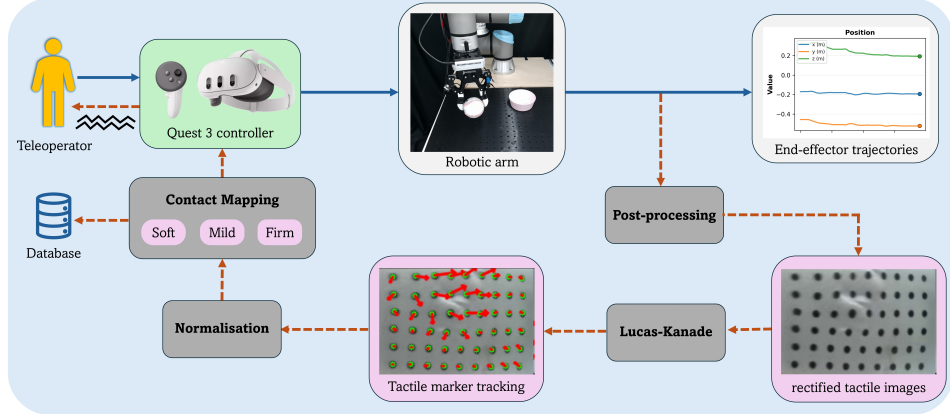} 
  \caption{Workflow for haptic feedback to the teleoperator from tactile marker tracking. Dashed lines show feedback transmission paths.      
}
  \label{fig_haptic}
\end{figure}

To close the perception–action loop during teleoperation,
\eqref{eq:tactile-motion-estimate} is converted into haptic feedback for the operator following the workflow in Fig.~\ref{fig_haptic}. The system normalizes \eqref{eq:tactile-motion-estimate} as
\begin{equation}
\widehat{m}_t =
\mathrm{clip}
\left(
\frac{m_t - m_{\min}}{m_{\max} - m_{\min}},
0, 1
\right).
\end{equation}
where $m_{\min}, m_{\max}$ are the minimum and maximum motion scores respectively, obtained prior to data collection. The normalized value is discretely mapped to vibration amplitude to indicate whether sufficient contact force is established. This discretization reduces sensitivity to sensor noise and provides clearer contact state transitions. The haptic signal is sent to the controller
as normalized left and right feedback values. Feedback is enabled only during active teleoperation, preventing misleading vibration during setup or reset.


\section{Experiments}
\label{sec:experiments}
Data collection follows a structured protocol to promote policy generalization. Object poses are randomized within a constrained workspace visible to the third-person camera, while the end-effector resets to a fixed initial position at the start of each episode. To increase scene diversity, one or two plausible clutter objects with randomized poses are added to each scene. Demonstration counts scale with task complexity, ranging from 30 to 60 episodes. For instance, 50--60 demonstrations were collected for longer-horizon or fluid-interaction tasks, such as bottle turning and board wiping, while 30 demonstrations were collected for simpler tasks. Each task includes two target states specified by the language instruction. This design improves language-action grounding by generating demonstrations with identical initial conditions but different instructions and trajectories, while also enabling generalization splits that test whether the policy conditions on language rather than converging to a single preferred outcome.

\subsection{Evaluation and Ablation Studies}
A subset of tasks from the HapTile dataset is selected for quantitative evaluation. The selected tasks are characterized by higher contact forces and richer tactile requirements than the broader dataset average, making them particularly suited to assess the value of visuotactile sensing over vision-only baselines. Four tasks for evaluation experiments are: \textbf{Turning a bottle upright}, \textbf{Wiping a whiteboard}, \textbf{Pouring liquid}, and \textbf{Peg insertion}. Detailed descriptions of each task are provided in Section \ref{sec:appendix_evaltasks}.

The dataset evaluation includes two key policy families. First, we use a Diffusion Policy~\cite{chi2025diffusion} baseline conditioned on visual, tactile, and proprioceptive features.
Second, we fine-tune the Vision-Language-Action (VLA) policy 
$\pi_0$~\cite{pi0}. Further information regarding deployment details is described in Section \ref{sec:appendix_evalpolicy}. We conduct ablation studies across four tasks using the learning policies, reporting success rate over 10 demonstrations in Table~\ref{tab_evaluation}. For each policy, we compare three modality configurations:\\
\begin{table}[t]
\centering
\caption{Success rates ($\mathbf{\%}$) of tasks using different learning policies and under varying modalities.}
\begin{tabular}{lccccccc}
\toprule
\multirow{2}{*}{\textbf{Tasks}} 
  & \multicolumn{3}{c}{\textbf{Diffusion Policy}} 
  & \multicolumn{3}{c}{\textbf{$\bm{\pi_{0}}$}} \\
\cmidrule(lr){2-4} \cmidrule(lr){5-7}
  & \textbf{V-only} & \textbf{V+T} & \textbf{V+TM} 
  & \textbf{V-only} & \textbf{V+T} & \textbf{V+TM} \\
\midrule
Turning a bottle upright 
  & 80 & 80 & \textbf{90} & 60 
  & 0 & 60 \\
Wiping a whiteboard 
  & 30 & 80 & 30 & 50 
  & 0 & \textbf{100} \\
Pouring liquid 
  & 50 & 50 & 20 & 30 
  & \textbf{60} & 0 \\
Peg insertion 
  & 40 & 40 & 0 & 0 
  & \textbf{90} & 40\\
\bottomrule
\end{tabular}
\label{tab_evaluation}
\end{table}
\textbf{Vision-only (V-only)} uses RGB camera observations from the third-person and wrist views alongside robot proprioception, with no tactile input. \\
\textbf{Vision with tactile sensing (V+T)} augments this with raw tactile images captured from sensors embedded in both the left and right fingers of the gripper.\\
\textbf{Vision with tactile sensing and marker tracking (V+TM)} further incorporates marker displacement features extracted from the tactile images, providing an explicit representation of contact geometry and slip.

\subsection{Results and Discussion}
Table~\ref{tab_evaluation} shows that adding tactile input generally improves or matches vision-only performance, although the optimal tactile representation depends on task physics. For peg insertion, $\pi_0$ improves from $0\%$ with vision alone to $90\%$ with raw tactile input (\textbf{V+T}), highlighting the importance of fine contact geometry. For whiteboard wiping, $\pi_0$ reaches $100\%$ under marker-tracked tactile input (\textbf{V+TM}) versus $50\%$ with vision alone, consistent with the role of shear and force cues. Similarly, \textbf{V+TM} achieves the best Diffusion Policy result ($90\%$) for turning a bottle upright.

The two tactile representations are complementary. Raw tactile images benefit tasks requiring precise contact localization, such as insertion and pouring, whereas marker tracking is more effective for force- and shear-dependent tasks such as wiping. However, tactile input does not consistently improve performance. Under Diffusion Policy with \textbf{V+TM}, wiping and pouring success rates decrease from $80\%$ to $30\%$ and from $50\%$ to $20\%$, respectively, suggesting over-reliance on high-dimensional tactile signals and motivating future work on adaptive fusion and modality gating.

The performance gap between Diffusion Policy and $\pi_0$ may reflect dataset scale. Diffusion Policy, being more compact, learns effectively from 50 demonstrations, while the larger pretrained $\pi_0$ struggles to align novel tactile inputs under limited fine-tuning data. Notably, $\pi_0$ performs more consistently with the structured \textbf{V+TM} representation.

\section{Limitations}
\label{sec:limitation}
\textbf{Scene Diversity and Generalization.} Data collection was conducted 
within a single laboratory scene. While objects and clutter span a broad range 
of daily-life activities reflecting environments such as kitchens and laundry 
rooms, the dataset lacks variation in scene layout, lighting, and background. 
Future work should incorporate multiple scene configurations and extend 
collection to in-the-wild settings to improve generalization.

\textbf{Simulation Data.} Scaling through physical collection alone is costly, 
requiring dedicated space, hardware, and supervision. Incorporating 
simulation data with realistic physics could enrich the dataset and facilitate 
large-scale benchmarking. Bridging the sim-to-real gap remains challenging, 
though recent advances make this an increasingly viable 
direction~\cite{zhao2026sim2real, liu2025dexscale, zhang2025vla}.

\textbf{Haptic Feedback Resolution.} The current haptic mechanism operates on 
force thresholds, distinguishing only two discrete force modes. While this 
binary signal provides meaningful contact feedback, it does not capture the 
continuous force variation characteristic of real manipulation. Graduated 
force levels would more faithfully reflect contact dynamics, particularly for 
tasks involving fragile, deformable, or surface-sensitive objects.


\section{Conclusion}
\label{sec:conclusion}
In this paper, we presented HapTile, a multimodal, contact-grounded manipulation dataset for learning and evaluating robot policies in everyday contact-rich tasks. HapTile provides language-conditioned demonstrations with synchronized vision, fingertip tactile sensing, robot proprioception, action trajectories, and haptic feedback collected through teleoperation on an accessible UR5e platform.

We described the dataset, tactile sensing pipeline, haptic feedback mechanism, and benchmark protocol, and provided baseline evaluations using Diffusion Policy and $\pi_0$. By combining diverse tasks, language conditioning, action trajectories, and embedded visuotactile sensing, HapTile provides a foundation for studying contact-aware policy learning and tactile vision-language-action models.

\clearpage

\bibliography{Refs}
\clearpage
\appendix
\section{Appendix}
\label{sec:appendix}
\subsection{Data Formatting}\label{sec:appendix_formatting}
Raw demonstrations are converted into a unified LeRobot/OpenPI-style format. Each episode contains metadata, structured state and action arrays, embedded visual streams, and language instructions.  Images are resized to a fixed resolution, and all state/action arrays are stored as \texttt{float32}. To avoid temporal leakage, we split the dataset at the episode level, so that all frames from a trajectory are assigned exclusively to the training, validation, or test set.

\subsection{Tactile Sensor Data Processing} \label{sec:appendix_tactile}

Since the two tactile sensors may be mounted with different viewing angles, we apply perspective rectification before using the tactile images for online processing. 
During calibration, four corner points of the valid sensing region are selected. 
A homography maps this quadrilateral region into a canonical rectangular image making the tactile coordinate frame consistent across sensors and collection sessions.

To reduce artificial tactile variation, we fix the camera exposure during collection instead of relying on automatic exposure. The system can store both the raw tactile images and rectified tactile images. Raw images are useful for auditing and reproducing the data pipeline, while rectified images are used for marker tracking, visualization, and policy inputs.

\subsection{Evaluation Tasks} \label{sec:appendix_evaltasks}
\textbf{Turning a bottle upright.} The robot approaches a water bottle lying horizontally on the table. The gripper must contact the bottle in the proximity of its center of mass to achieve a stable grasp, after which the robot executes a reorientation trajectory that places the bottle base on the table surface and releases it in a stable upright configuration. The task demands precise force regulation to avoid slipping on the curved surface during reorientation.

\textbf{Wiping a whiteboard.} This task follows a protocol used as a standard benchmark in several prior studies~\cite{elliott2017learning, zhu2020robosuite, huang2025tactile, lew2023robotic}. The gripper approaches and grasps a sponge, then transports it to a region of the whiteboard marked with ink. The robot must apply adequate normal force against the board surface while executing a lateral wiping motion along the marked area to erase it. The task requires sustained contact force control, captured by tactile image variations, throughout the wiping trajectory.

\textbf{Pouring liquid.} A bottle of liquid cleanser is positioned upright on the table. The gripper grasps the bottle from the side and tilts it beyond horizontal, inclining the opening downward past parallel to the table surface, aligned with one of two target bowls specified by the language instruction. Successful execution requires a smooth and controlled tilt trajectory to avoid spillage during the pour.

\textbf{Peg insertion.} A D-shaped peg is placed on the table near a target hole. The gripper grasps the peg, aligns it with the hole, and inserts it until fully seated. Precise alignment and controlled insertion force are required, as misalignment necessitates readjustment before successful insertion.

\subsection{Policy Training and Deployment} \label{sec:appendix_evalpolicy}
We use Low-Rank Adaptation (LoRA) fine-tuning~\cite{hu2022lora} for memory-efficient adaptation.
Before training, we compute normalization statistics for states and actions. At time $t$, the policy receives a multimodal observation
\begin{equation}
    o_t =
    \{I_t^{base}, I_t^{wrist}, I_t^{left\_tactile},
    I_t^{right\_tactile}, s_t, l\},
\end{equation}
where $I_t^{base}$ is the third-person image, $I_t^{wrist}$ is the wrist image, $I_t^{left\_tactile}$ and $I_t^{right\_tactile}$ are tactile images, $s_t$ is robot proprioception, and $l$ is the language instruction. The policy predicts an action chunk
\begin{equation}
    a_{t:t+H-1}\sim\pi_\theta(o_{\leq t}).
\end{equation}
At deployment time, we execute the first part of the predicted chunk and then replan with the latest observation.
During real-robot deployment, a GPU-side policy server returns action chunks over WebSocket, while the robot-side controller keeps all safety checks, including action clipping, maximum step limits, workspace bounds, and emergency stop handling.


\newpage

\end{document}